\pdfoutput=1

\documentclass{llncs}

\usepackage{makeidx}

\usepackage[utf8]{inputenc}
\usepackage{graphicx}
\usepackage{url}
\usepackage[misc]{ifsym}

\begin{document}

\mainmatter

\title{Towards co-evolution of fitness predictors and Deep Neural Networks}

\author{Włodzimierz Funika\inst{1,2} \and Paweł Koperek\inst{2}}

\institute{
    AGH, ACC CYFRONET AGH, ul. Nawojki 11, 30-950, Kraków, Poland
    \and
    AGH, Faculty of Computer Science, Electronics and Telecommunication, Dept. of Computer Science, al. Mickiewicza 30, 30-059, Kraków, Poland
    \begin{center}
        email:{funika@agh.edu.pl, koperek@icsr.agh.edu.pl}
    \end{center}
}

\maketitle

\begin{abstract}
 
    Deep neural networks proved to be a very useful and powerful tool with many practical applications. They especially excel at learning from large data sets with labeled samples. However, in order to achieve good learning results, the network architecture has to be carefully designed. Creating an optimal topology requires a lot of experience and knowledge. Unfortunately there are no practically applicable algorithms which could help in this situation. Using an evolutionary process to develop new network topologies might solve this problem. The limiting factor in this case is the speed of evaluation of a single specimen (a single network architecture), which includes learning based on the whole large dataset. In this paper we propose to overcome this problem by using a fitness prediction technique: use subsets of the original training set to conduct the training process and use its results as an approximation of specimen's fitness. We discuss the feasibility of this approach in context of the desired fitness predictor features and analyze whether subsets obtained in an evolutionary process can be used to estimate the fitness of the network topology. Finally we draw conclusions from our experiments and outline plans for future work.

\end{abstract}

\section{Introduction}

Deep neural networks (DNN) are a very powerful machine learning technique. They have numerous practical applications, with state-of-the-art performance reported in several domains, ranging from visual object recognition (\cite{Krizhevsky2012}), through text processing (\cite{Bengio2003}, \cite{Collobert2008}) to speech recognition (\cite{Dahl2012}, \cite{Hinton2012}). 

Neural network models are especially well suited to tackle problems with available large data sets of labeled samples. Model capacity can be easily increased by adding more units (neurons) in layers or by adding more layers. Unfortunately, choosing the correct architecture is not straight-forward. Even given a set of layers with optimal types and sizes, the learning process may ultimately fail: the resulting model can be under-performing in terms of accuracy or can be overfitted. To combat these kind of problems, a number of techniques was developed: L1/L2 regularization \cite{Ng2004}, dropout/dropconnect (\cite{Srivastava2014}, \cite{Wan2013}), early stopping, pre-training \cite{Courville2010}, adaptive learning rate \cite{Cho2011} etc. Each of them has own limitations and has to be used in specific context to actually improve the results. Building a well-performing model requires a lot of experience and performing some experiments with the actually analyzed dataset.

Using the automated methods for creating deep neural network models would greatly improve their quality and speed up creating innovative structures. As demonstrated by Koza in \cite{Koza2010}, the use of evolutionary algorithms can provide complex problem solutions, whose quality is comparable to those created by a human. There are examples of successful application of this approach to the benchmark problems (\cite{Jha2016}, \cite{MohamedBenAli2008}) and real world challenges (\cite{Khan2016}). The factor which limits the usability of such an approach for DNN is the time of training of the network - the time that it takes to evaluate the model. The mentioned models were relatively simple (hundreds of neurons) and had limited training datasets (thousands of samples). The Deep Neural Networks are much more complex. Additionally, increasing the scale of deep learning in respect to both training examples and the number of parameters, is recognized as the main factor which improves the quality of results of the learning process. Conducting the training requires a lot of processing resources. In recent years there were many reports of successfully using GPUs (\cite{Krizhevsky2012}, \cite{Raina2009}, \cite{Dahl2012}) or powerful large-scale clusters \cite{Dean2012}, \cite{Le2011} to scale up training and inference algorithms. Unfortunately, due to the increase of data sets size, the speed of research still suffered. It is still necessary to wait hours to days or weeks in order to learn whether a chosen topology combined with specific learning parameters provide optimal results. 

Evolutionary methods would become applicable if only the evaluation time could be reduced. One method which helps in this situation is the co-evolution of so called \emph{fitness predictors} \cite{Schmidt2006}, \cite{Schmidt2008}, \cite{Funika2014}. It assumes that a low-cost heuristic can be used to compare individuals in the population, instead of performing time consuming evaluation over the full dataset. Since the processing time is greatly reduced, evolutionary algorithms become feasible again. In this paper we present our approach to the evaluation of using subsets of the training set as fitness predictors for Deep Neural Networks. We analyzed the properties of such a solution and verified the hypothesis with some experiments using the MNIST dataset \cite{lecun-mnisthandwrittendigit-2010}. 

The paper is structured as follows: the next section describes the background and related work. Next we discuss the feasibility of using subsets of training set as fitness predictors, present sample results of fitness predictor evolution and analyze the potential problems. Finally we conclude the results and provide directions for further research.

\section{Background and related work}

In this section we present the related work which sets the foundation for our research.

\subsection{Deep Neural Networks}

In the standard approach a neural network consists of many simple connected processing units called neurons. Each of them produces a real number, which is the result of an activation function applied to sum of inputs multiplied by corresponding weights. The neurons are grouped into so-called \emph{layers}. The first one (the \emph{input layer}) gets activated by sensors observing some environment. Further layers are formed by connecting outputs of one of them to inputs of another. Finally, some of the neurons - typically forming the last (\emph{output}) layer, might influence the environment by triggering some actions or can be used as a model of some phenomenon. 

\emph{Learning} in neural networks is a process of finding the weight values that make the network exhibit a \emph{desired} behavior. Depending on the problem and environment complexity, such behavior might require many computational stages. Each of them transforms, usually in a non-linear way the aggregated activation of the network. In \emph{Deep Neural Networks} there are \emph{many} such stages. 

\emph{Shallow} networks have been known for a long period of time, with the earliest works published in 1940s (\cite{McCulloch1943}). Models with successive layers of neurons were introduced a bit later \cite{rosenblatt1958}, \cite{Narendra:74}, however it took time to develop practical, efficient learning methods \cite{Werbos:81sensitivity}. Using this approach was not popular at first, because of practical problems: the learning algorithm was not proven to reliably find a nearly optimal global set of weights for complex problems in a reasonable amount of time. It required many subsequent improvements, like introducing convolutional and sub-sampling layers \cite{Fukushima:1979neocognitron}, using GPUs \cite{chellapilla:2006b}, \cite{ranzato:2006}, max-polling \cite{weng1992}, L1/L2 regularization (\cite{Ng2004}), dropout and dropconnect (\cite{Srivastava2014}, \cite{Wan2013}), early stopping, pre-training \cite{Courville2010}, adaptive learning rate (\cite{Cho2011}), to successfully apply networks with many hidden layers (\emph{deep neural networks}) to practical problems like visual object recognition (\cite{Krizhevsky2012}), text processing (\cite{Bengio2003}, \cite{Collobert2008}) or speech recognition (\cite{Dahl2012}, \cite{Hinton2012}). Nowadays, thanks to successes in international competitions (\cite{seung2009}, \cite{stallkamp:11}, \cite{stallkamp:12}, \cite{ciresan:2010deepbig_arxiv}, \cite{ciresan:2011ijcnn}), neural networks gained wide-spread attention and are a very dynamic field of research.

\subsection{Evolution Algorithms and Neural Networks}

Using the numerous layer types, techniques and learning algorithms introduces an additional layer of parameters to the machine learning system: the \emph{hyper-parameters} (as opposed to the parameters of the model - \emph{weights} of the neurons inputs). Choosing them properly has a great impact on the overall performance and accuracy of a specific network topology. Together with the variety of network topologies, it makes the task of designing, learning and applying deep neural network very complicated.  

In shallow neural networks, the neuroevolution - artificial evolution of neural networks using genetic algorithms, has shown a great promise to improve the situation. Evolution have been be applied in different scenarios, with majority research done in the following three scenarios: evolving the connection weights values, evolving the network topology or evolving both. 

In the first case the evolution is effectively replacing the back-propagation algorithm \cite{Montana1989}. It promises to overcome the drawbacks of gradient-descent based methods like trapping in local minima or inability to find a global minimum of a nondifferentiable functions. In practice however, those problems are not commonly observed and are outweighed by advantages: speed and scalability. Furthermore, this approach still requires creating and tuning the network architecture.

Evolving only the network topologies \cite{Mandischer1995}, \cite{Kitano1990} is the second option. In this approach, a topology can be created using different strategies: extending a very basic, minimal network with new neurons and connections (\emph{growing}) \cite{Stanley2002}, \cite{Xinjian2010} or starting with a big network and gradually removing its elements (pruning) \cite{Siebel2009}. During the evaluation, the networks are trained and tested against a separate test set. Topologies obtained in this way were reported to provide better generalisations. Processing them should also be more performant, as they contain only the necessary layers and neurons, thus limit the amount of the needed computations.

The algorithms which create the third group are often referred to as \emph{Topology and Weight Evolving Artificial Neural Network} (TWEANN) algorithms. The most widely known ones include: Neuroevolution of Augumenting Topologies (NEAT) \cite{Stanley2002} and HyperNEAT \cite{Stanley2009}, Cartesian Genetic Programming Artificial Network (CGPANN) algorithm \cite{Khan2010}, GeNeralized Acquisition of Recurrent Links (GNARL) \cite{Angeline1994}. Evolving the topology and weights together is reported to provide better results than any of them alone \cite{XinYao1999}. Unfortunately, due to technical limitations, this method has been applied only to benchmark problems like the single pole balancing \cite{Khan2010}. Applying this method to more complex use-cases like visual pattern recognition has not been widely investigated yet.

The area of neuroevolution in deep neural networks is not explored to such an extent yet. In \cite{Verbancsics2013} HyperNEAT is used to train a neural network which learns to classify images from the classic MNIST dataset (\cite{lecun-mnisthandwrittendigit-2010}. In this approach, the topology of the network was pre-defined and evolutionary algorithm was used to find weight values. Experiments included two scenarios: 

\begin{itemize}
    \item finding weights across all layers
    \item finding weights for feature extracting layers of a convolutional network and combining them with a traditional neural network which was trained with back-propagation
\end{itemize}

In the best configuration, that approach achieved 92.1\% accuracy which is subpar to the results obtained with gradient-descent methods. 

In \cite{David2014} a genetic algorithm is used together with back-propagation to conduct training of a neural network. For each layer of the network there are multiple sets of weights. In each iteration this set is evolved and the most fit individual (set of weights with the smallest root mean squared error over training samples) is chosen and further tuned with back-propagation. The authors describe how this approach was applied to the problem of classification of images from the MNIST dataset and report achieving a classification test error of only 1,44\%. 

As recognized in \cite{Tirumala2014}, in the context of deep neural networks, the main objective for using the evolutionary algorithms was to improve the learning mechanisms. Unfortunately, this approach, at least at this stage of development, can not be used to replace the gradient-based methods. On the other hand, the topology evolution in deep neural networks was not explored extensively yet. The use of traditional algorithms, like NEAT and GNARL, is limited by the time which has to be spent on training and testing a single individual. We believe that to unlock the potential of those algorithms, research has to focus on making the evaluation time shortest as possible.

\subsection{Co-evolution of fitness predictors}

Co-evolution is a kind of evolutionary algorithm where one individual within the same or a separate population, is used to determine the relative ranking between other individuals \cite{Bongard2005}, \cite{Olsson2001}. In other words, whether individual A is inferior or superior to individual B may depend of a third individual C rather than on some external fitness metric which would provide an absolute ranking. There is a number of different forms of co-evolution: antagonistic (e.g. predator-prey), cooperative (e.g. symbiosis) or nonsymmetric systems (e.g. host-parasite or teacher-learner). Fitness in the context of co-evolution has two notions: \emph{objective} and \emph{subjective}. The former is the well defined absolute ordering metric used in classical evolutionary algorithms. The latter is defined by the coevolving individuals and may be only weakly correlated with the objective fitness. 

One of the major limiting factors in the evolutional computations is the time of single individual evaluation. One approach to tackle this problem is to use the fitness modelling \cite{Jin2005} techniques, which attempt to approximate the exact fitness by using a model or coarse simulation of analyzed system. In the context of evolutionary computations the major techniques include:

\begin{itemize}
\item \emph{fitness inheritance} - fitness values are transferred from parents to children during crossover 
\item \emph{fitness imitation} - individuals are clustered by using a distance metric. The central individual of each cluster is evaluated in full and the resulting fitness is assigned to all elements of the cluster. 
\item \emph{partial evaluation} - fitness for some individuals is calculated exactly, others are inherited or modeled.
\end{itemize}

A chosen modelling method can be incorporated into the evolutionary process in many ways, e.g. to initialize the population, guide the crossover and mutation or replace fitness evaluations. Such an approach has numerous advantages: it reduces the evaluation cost and frequency while maintaining evolutionary progress, destabilizes local optima, helps avoiding bloated solutions and it can be applied in a situation where no explicit fitness function exists. Unfortunately there are also many challenges which come with using fitness approximation, like choosing the correct model, training it properly or dealing with a loss of fitness accuracy.

Both ideas (co-evolution and fitness approximation) have been combined and applied successfully to a field, which also suffers from long evaluation times on big data sets - symbolic regression \cite{Schmidt2009}, \cite{Schmidt2006}, \cite{Schmidt2008}. In this approach, the fitness prediction technique was used. It replaces exact fitness evaluations with a light-weight approximation which adapts together with the solution population. To achieve that, a population of so called \emph{fitness predictors} is co-evolved together with the problem solution population. Their objective is to maximize prediction accuracy. The best of the predictors is used to evaluate the solutions to the original problem. Fitness predictors in the case of symbolic regression were encoded as a small subset of the full training data set. This allowed to dramatically speed up computations, improved the quality: reduced bloat and increased the fitness values of solutions of the original problem.

\section{Subsets of training set as fitness predictors}

There are many ways to represent the fitness predictor and how it is evolved and used. It is very important to choose an appropriate form correctly. The chosen predictors are used for all fitness evaluations within evolutionary algorithm iteration, hence they influence the direction of development of the main population. As stated in \cite{Schmidt2006}, the following constraints have to be met by fitness predictors:

\begin{enumerate}

    \item They have to be able to approximate the fitness of a candidate solutions
    \item They have to be processed significantly faster than the exact fitness calculation
    \item They have to differentiate the fitness between a pair of individuals from a given population.

\end{enumerate}

Fitness prediction can be conducted with use of different estimation methods and each of them will impose a different representation of a single predictor. One might use e.g. a decision tree and this would force use a decision tree representation which could be subjected to evolution. In our case, where we want to estimate the fitness of Deep Neural Networks, we propose to conduct the estimation by training and testing with an unchanged algorithm but by using each time a different subset of the full training data set. This allows to represent a single fitness predictor in a very simple way: as an array of indexes of the full data set. This approach have a number of advantages:

\begin{itemize}

    \item It is easy to tune the speed vs accuracy of approximation with the size of the subset
    \item The representation is very simple - the implementation is less error prone
    \item The evaluation of a single individual is basically the same procedure as training the neural network. It is possible to carry out experiments with different hyper-parameters of the network.

\end{itemize}

The major disadvantage of our approach is the risk of over-fitting the model because of using relatively small training sets. Given the complexity of a typical Deep Neural Network, where millions of parameters are not uncommon, over-fitting is almost certain. However, we believe that this effect can be limited by using certain techniques e.g. the dropout layers. 

We evaluated our idea in the environment described in \cite{LeCun1998}, where a Convolutional Neural Network was used to recognize hand written digits from \cite{lecun-mnisthandwrittendigit-2010}. First we created a topology based on the LeNet5 (Table \ref{table:topology}) using the Torch7 framework \cite{Collobert2011}. Initially we used the full original dataset consisting of 60000 training samples. Training was conducted using the Stochastic Gradient Descent method with mini-batches. With this setup, we achieved an accuracy of 99,21\% on the provided validation test set. The learning parameters are listed in Table \ref{table:dnn_learning_parameters}. Such a result is on par with the original model. 

\begin{table}
    \center
    \begin{tabular}{|c|c|c|c|}

%        \hline
%        No. & Layer Type & Layer Size & Number of parameters & Output size \\
%        \hline
%         1. & Convolution (ReLU \cite{Glorot2011} activation) & 32 planes ($5 \times 5$ receptive fields) & 832 & $32 \times 24 \times 24$ \\ 
%         2. & Max-Pooling & 32 planes aggregating $3 \times 3$ squares & 0 & $32 \times 8 \times 8$ \\
%         3. & Convolution (ReLU activation) & $32 \times 64$ planes ($ 5 \times 5 $ receptive field) & 53248 & $64 \times 4 \times 4$ \\
%         4. & Max-Pooling & 64 planes aggregating $2 \times 2$ squares & 0 & $64 \times 2 \times 2$ \\
%         5. & Fully connected & 200 neurons & 25600 & 200 \\
%         6. & Fully connected & 10 neurons & 2000 & 10 \\
%         7. & LogSoftMax & 1 neuron & 0 & 1 \\
%        \hline

        \hline
        No. & Layer Type & Number of parameters & Output size \\
        \hline
         1. & Convolution (ReLU \cite{Glorot2011} activation) & 832 & $32 \times 24 \times 24$ \\ 
         2. & Max-Pooling & 0 & $32 \times 8 \times 8$ \\
         3. & Convolution (ReLU activation) & 53248 & $64 \times 4 \times 4$ \\
         4. & Max-Pooling & 0 & $64 \times 2 \times 2$ \\
         5. & Fully connected & 25600 & 200 \\
         6. & Fully connected & 2000 & 10 \\
         7. & LogSoftMax & 0 & 1 \\
        \hline
    \end{tabular}
    \caption{Topology of the network used in experiments: layer types, their size and number of parameters a layer introduces in the model (includes bias).}
    \label{table:topology}
    \label{}
\end{table}

\begin{table}

    \center
    \begin{tabular}{|c|c|}
        \hline
        Parameter & Value \\
        \hline
        Learning rate & 0.1 \\
        Learning rate decay & 0.00001 \\
        Mini-batch size & 128 \\  
        Learning epochs & 20 \\
        L1/L2 coefficients & 0 \\
        Momentum & 0 \\
        \hline
    \end{tabular}

    \caption{Neural network training parameters used in experiments}
    \label{table:dnn_learning_parameters}

\end{table}

\subsection{Approximation of fitness}

First, we evaluated whether the proposed fitness predictors can accurately predict the fitness of a given candidate - a deep neural network. Using a subset of the training data set has an obvious advantage: in the worst case scenario the model will be able to recognize at least a part of the full data set. However, the samples should not be picked randomly as the quality of the chosen subset has a great influence on the result of learning. To quantify how big the impact of that effect is, we attempted to measure the difference in the prediction accuracy, when random and pre-selected subsets are used. 

\begin{figure}[t]
    \centering
    \includegraphics[width=0.75\textwidth]{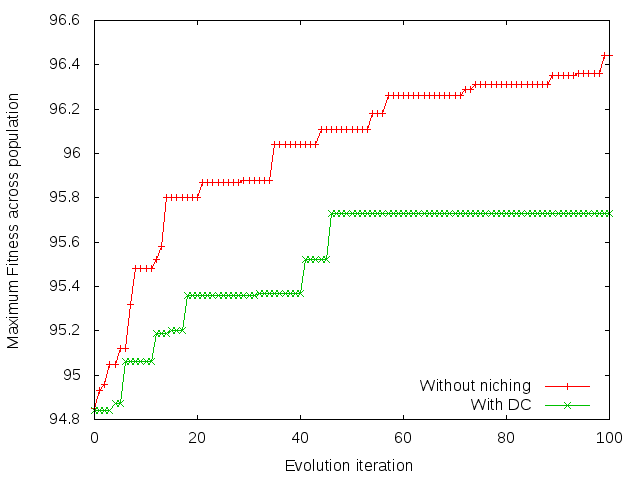}
    \caption{Maximum fitness in the population in subsequent iterations of evolution with Deterministic Crowding (DC) niching technique and without it.}
    \label{figure:fitness_niching}
\end{figure}

To find the subset which has the best training performance we implemented a genetic algorithm which evolved a population of potential fitness predictors with a fixed size. The mutation operation was implemented as a simple change of a single training sample with another, randomly chosen one. For simplicity we chose the one-point variant of cross-over. We experimented with different values of the cross-over point location (fixed and random), however we did not observe any differences in that algorithm convergence speed. We also investigated using a niching technique (Deterministic Crowding \cite{Mengshoel2008}) versus simply choosing the best individuals from a combined children and parents population. The latter, simpler approach turned out to converge faster. The fitness in subsequent iterations with and without niching is presented in Fig. \ref{figure:fitness_niching}. The parameters of the evolutionary algorithm are listed in Table \ref{table:evolution_parameters}.

\begin{table}

    \center
    \begin{tabular}{|c|c|}
        \hline
        Parameter & Value \\
        \hline
        Population size & 128 \\
        Evolution iterations & 100 \\
        Cross-over probability & 75\% \\
        Mutation probability & 1\% \\        
        Number of iterations & 100 \\
        Cross-over variant & Single point \\
        \hline
    \end{tabular}

    \caption{Parameters of the evolutionary algorithm, which rendered the best results in conducted experiments.}
    \label{table:evolution_parameters}

\end{table}

Each predictor was evaluated using the same procedure:

\begin{enumerate}

    \item A new instance of the neural network was instantiated.
    \item The network was trained using the samples specified by the genotype of a given individual. We used the same training method as described earlier.
    \item Network accuracy was evaluated using the full test dataset.

\end{enumerate}

The results of evolution with different sizes can be compared reliably thanks to the use of the same validation set for each fitness predictor size.

\begin{figure}[t]
    \centering
    \includegraphics[width=0.75\textwidth]{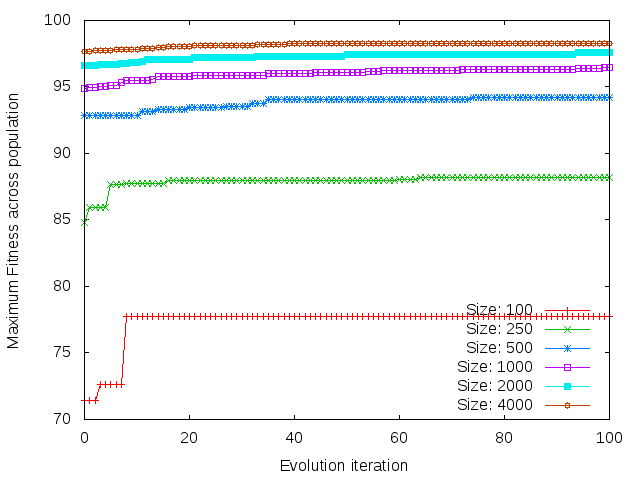}
    \caption{Maximum fitness in the population in subsequent iterations for increasing dataset size (min = 100, max = 4000).}
    \label{figure:fitness_sizes}
\end{figure}

The maximum accuracy across the population of the subsets in the subsequent evolution iterations is presented in Fig. \ref{figure:fitness_sizes}. It is clear that with increasing the size of the fitness predictor, the accuracy of the trained neural network is higher. Regardless of the size, the evolution was able to improve the prediction results. The effect is weaker for bigger fitness predictors, however it is clear that using just a random subset might result in poor performance even if the model has an optimal architecture. The differences between the worst and best individuals are presented in Table \ref{table:fp_fitness_diff}.

\begin{table}

    \center
    \begin{tabular}{|c|c|c|c|c|}
        \hline
        Size & Min & Max & Max - Min & Full dataset - Max\\
        \hline
        100 & 9,86\% & 77,76\% & 67,9\% & 21,45\% \\
        250 & 50,77\% & 88,21\% & 37,44\% & 11,00\% \\
        500 & 44,89\% & 94,24\% & 49,35\% & 4,95\% \\
        1000 & 84,86\% & 96,44\% & 11,58\% & 2,77\% \\
        2000 & 94,23\% & 97,56\% & 3,33\% & 1,65\% \\
        4000 & 95,94\% & 98,28\% & 2,34\% & 0,93\% \\
        \hline
    \end{tabular}

    \caption{Differences in the maximal and minimal accuracy and reference (full dataset) and max accuracy of models trained with use of a subset of a given size.}
    \label{table:fp_fitness_diff}

\end{table}

Based on these results we conclude that training with only a subset of available training samples can be used to approximate the prediction accuracy of the full training dataset. However, the subset has to be chosen very carefully. It is possible to pick 250 and 1000 element datasets which will result in training a model with a similar recognition accuracy.

\subsection{Time of processing}

Cutting down the time of evaluating a single individual (neural network) is the main objective of this research. In this section we compare the time of training our deep neural network with use of the full training set and fitness predictors of different sizes.

Table \ref{table:fp_eval_time} presents average times of a single iteration (\emph{epoch}) of learning for each fitness predictor size. We used the same test set to evaluate the model's accuracy in each case. Therefore in the table we exclude the evaluation time completely as it is invariant to the input size.

\begin{table}

    \center
    \begin{tabular}{|c|c|c|}
        \hline
        Fitness Predictor Size & Average training time & Standard deviation \\
        \hline
        100 & 100.37 & 9.94 \\
        250 & 245.12 & 12.15 \\
        500 & 492.63 & 17.26 \\
        1000 & 981.26 & 29.67 \\
        2000 & 1923.69 & 46.93 \\
        4000 & 3890.54 & 87.64 \\
        60000 & 57698.10 & 369.26 \\
        \hline
    \end{tabular}

    \caption{Average time of learning using a fitness predictor of a specified size (in milliseconds). Computed over 100 runs for each fitness predictor size.}
    \label{table:fp_eval_time}

\end{table}
 
The times were measured on a Intel Xeon processor with 8 cores. For each fitness predictor size a single epoch of training was executed 100 times. 

The results present a linear relationship between the data set size and training time. The time is extended by roughly 1ms per each sample. It is clear that by limiting the number of samples a considerable amount of time can be saved. By connecting this relationship to the size-prediction accuracy dependence (Table \ref{table:fp_fitness_diff}) we obtain a tool which enables tuning the accuracy of trained neural network (the \emph{approximated fitness}) to a level, where the processing time is acceptable. The only parameter which needs to change is the size of the fitness predictor - more elements result in improved accuracy at the cost of longer training time.

We acknowledge that the presented timing results shouldn't be treated as a benchmark of Torch performance: they are related only to a single dataset and the system was not fine-tuned to reach peak performance.

\subsection{Differentiating the fitness between individuals}

One important feature of computing a fitness score, is the possibility to compare different individuals according to specific criteria. In our case this would mean comparing between different neural network architectures in the context of their capability to generalize knowledge from the provided training samples. Using only a subset of the original training set increases the pressure to generalize even more. All models start with the same subset of the training data set and are evaluated with use of the same test set. This means they are exposed in the same way to any flaws of fitness predictors, e.g. the distribution of samples is extremely biased towards one of the classification categories, in case of MNIST dataset, one of the digits. The fitness score might not create an \emph{objective} (absolute) ordering of elements, but will provide a reliable \emph{subjective} (relative) comparison within a population. Given that the fitness predictors improve over time, this enables to improve the general population of the network. 

\section{Conclusions and further work}

In this paper we formulated and discussed the concept of using subsets of training data sets as fitness predictors for deep neural networks. We analyzed whether this form meets the fitness predictor criteria. Further we presented the results of an experiment which attempted to improve the quality of fitness predictors with use of an evolutionary algorithm. The classification accuracy of networks, after training with fitness predictors of different sizes, combined with the time of computations, showed that the accuracy can be traded off for shorter time of processing.

Those results prove that the proposed approach to fitness prediction of deep neural networks is a viable alternative to the evaluation by training and verifying over a complete training and test sets. This doesn't replace the technique of using largest available datasets to achieve high quality results with the optimal network structure. However, it enables optimization of time and resources where many instances of Deep Neural Networks are compared to each other, e.g. in the case of using the evolutionary algorithms to find the optimal number of layers.

Furthermore, the subsets of training set found by the evolutionary algorithm, can be reused by other researchers to speed up experimentation with different network structures. This will enable further improvements in the area of deep learning. 

We plan to continue this research by implementing a co-evolutionary algorithm which will adjust the architecture of a deep neural network to a specific problem. Fitness predictors will be used to significantly cut down the evaluation time of a single network. This will optimize the resource usage and allow much more models to be evaluated. We also want to explore the methods of improving the evolution process by introducing another population which could be used to tune the values of hyper-parameters.

\section*{\ackname} We would like to thank the PL Grid project for providing computational resources to carry out computational experiments.

\section*{Funding} This research is partly supported by AGH grant no. 11.11.230.124. 

\bibliographystyle{unsrt}
\bibliography{towards_coevolution_dnn}
\end{document}